\documentclass[lettersize,journal]{IEEEtran}

\usepackage[normalem]{ulem}
\usepackage{stfloats}
\usepackage{bm}

\usepackage{tabularray}
\usepackage{caption}
\usepackage{subcaption}
\usepackage{amsmath,amsfonts}
\usepackage{array}
\usepackage{textcomp}
\usepackage{stfloats}

\usepackage{url}
\usepackage{verbatim}
\usepackage{graphicx}
\usepackage{cite}
\usepackage{multirow}
\usepackage{booktabs}

\usepackage{xcolor}

\usepackage{xspace}
\newcommand{\name}[0]{PAKT\xspace}

\usepackage{amssymb}

\usepackage{bbding}
\makeatletter
\DeclareRobustCommand\onedot{\futurelet\@let@token\@onedot}
\def\@onedot{\ifx\@let@token.\else.\null\fi\xspace}

\usepackage{scalerel}
\usepackage{tikz}
\usetikzlibrary{svg.path}
\definecolor{orcidlogocol}{HTML}{A6CE39}
\tikzset{
    orcidlogo/.pic={
        \fill[orcidlogocol] svg{M256,128c0,70.7-57.3,128-128,128C57.3,256,0,198.7,0,128C0,57.3,57.3,0,128,0C198.7,0,256,57.3,256,128z};
        \fill[white] svg{M86.3,186.2H70.9V79.1h15.4v48.4V186.2z}
        svg{M108.9,79.1h41.6c39.6,0,57,28.3,57,53.6c0,27.5-21.5,53.6-56.8,53.6h-41.8V79.1z M124.3,172.4h24.5c34.9,0,42.9-26.5,42.9-39.7c0-21.5-13.7-39.7-43.7-39.7h-23.7V172.4z}
        svg{M88.7,56.8c0,5.5-4.5,10.1-10.1,10.1c-5.6,0-10.1-4.6-10.1-10.1c0-5.6,4.5-10.1,10.1-10.1C84.2,46.7,88.7,51.3,88.7,56.8z};
    }
}
\newcommand\orcidicon[1]{\href{https://orcid.org/#1}{\mbox{\scalerel*{
                \begin{tikzpicture}[yscale=-1,transform shape]
                \pic{orcidlogo};
                \end{tikzpicture}
            }{|}}}}
\usepackage{hyperref}
\hypersetup{hidelinks,
	colorlinks=true,
	allcolors=black,
	pdfstartview=Fit,
	breaklinks=true}

\usepackage{algorithmicx}
\usepackage{algpseudocode}  
\usepackage{algorithm}  

\usepackage[numbers, sort&compress]{natbib}

\begin{document}


\title{Disentangling Knowledge States with Ability and Proficiency Modeling for Knowledge Tracing}




\author{
Duantengchuan Li\textsuperscript{\orcidicon{0000-0003-2902-7365}},
Yingqian Bi\textsuperscript{\orcidicon{0009-0003-3672-3470}},
Jinsong Chen\textsuperscript{\orcidicon{0000-0001-7588-6713}},
Rui Zhang\textsuperscript{\orcidicon{000-0001-9418-0863}},
Mingwen Tong\textsuperscript{\orcidicon{0000-0002-9524-7983}}

\IEEEcompsocitemizethanks{
\IEEEcompsocthanksitem Duantengchuan Li and Rui Zhang are with the School of the Information Management, Wuhan University, Wuhan 430072, China.
\IEEEcompsocthanksitem Yingqian Bi is with the School of Computer Science and Artificial Intelligence, Hubei University of Technology, Wuhan 430068, China; School of the Information Management, Wuhan University, Wuhan 430072, China
\IEEEcompsocthanksitem Jinsong Chen and Mingwen Tong are with the Faculty of Artificial Intelligence in Education, Central China Normal University, Wuhan 430074, China.

}
\thanks{The first two authors contribute equally.}

\thanks{(Corresponding author: Jinsong Chen and Rui Zhang. E-mail: guangnianchenai@ccnu.edu.cn and ruizhang8633@gmail.com.)}
}
\markboth{Journal of \LaTeX\ Class Files,~Vol.~??, No.~??, ??~202?}%
{Shell \MakeLowercase{\textit{et al.}}: A Sample Article Using IEEEtran.cls for IEEE Journals}


\maketitle
\begin{abstract}

Knowledge tracing (KT) aims to predict students' future performance by modeling their evolving knowledge states from historical interactions. 
Existing KT methods usually treat the raw interaction sequence as a unified behavioral process, overlooking the phase-specific nature of learning behaviors. 
Our preliminary observations show that students are more likely to correctly answer previously failed knowledge concepts after sufficient practice, suggesting a transition from ability-building to proficiency-oriented learning. Motivated by this, we propose Phase-Aware Knowledge Tracing (\name), a KT framework that decomposes student interactions into ability and proficiency phases based on the tailored decomposition mechanism. 
To effectively exploit the decomposed sequences, we design a multi-branch Transformer with a type-aware readout module to jointly capture phase-specific and holistic knowledge states.
We further provide a causal analysis to reveal the confounding bias caused by entangling complex learning behaviors in phase-agnostic KT models. 
Extensive experiments on six public benchmarks demonstrate that our method consistently outperforms representative baselines, with a maximum AUC gain of 1.33\% and an average gain of 0.82\%.

\end{abstract}


\section{Introduction}
With the development of educational data mining and learning analytics technologies, Intelligent Tutoring Systems (ITS) have gained significant attention\cite{aiapp}.
An ITS is a computer-based system that delivers personalized instruction and adaptive feedback to students. 
By modeling the strengths and weaknesses of a student, ITS provides adaptive problem selection, learning performance prediction, and timely feedback to facilitate effective learning outcomes.
A central challenge of ITS lies in accurately modeling the evolving knowledge state of individual students.
To address this, Knowledge Tracing (KT), which aims to estimate a student's mastery of knowledge concepts from their historical interaction sequences, has emerged as a fundamental task in ITS.

Early KT models are grounded in probabilistic modeling psychometric theory, including Hidden Markov Models (HMMs) and Item Response Theory (IRT), for student performance prediction~\cite{bkt,irt}.
DKT\cite{dkt} first introduces deep learning to the KT task  with recurrent neural networks (RNNs), substantially outperforming prior probabilistic approaches.
Since each student's interaction history is naturally sequential, deep learning architectures for sequential data have been widely applied to the KT task~\cite{deepirt, sakt, sinkt}.
Sequential KT research has converged on three primary modeling approaches.
The first consists of deep sequential models, such as the RNN-based DKT\cite{dkt} and the Transformer-based SAINT~\cite{saint,transformer}, which capture temporal dependencies in student interaction sequences to estimate evolving knowledge states.
The second covers structure-enhanced KT models, which incorporate auxiliary structures, such as external memory in DKVMN\cite{dkvmn} and concept prerequisite graphs in GKT\cite{gkt} and GRKT\cite{grkt}.
The third comprises interpretable KT models, such as EKT\cite{ekt}, XKT~\cite{xkt,lstm}, and DisenKT\cite{disenkt}, which enhance transparency by incorporating educational theories or learning disentangled representations of student knowledge states~\cite{causalKT, cikt, mikt,rckt}.


Despite the remarkable progress achieved by existing KT methods~\cite{tcnet,simplekt}, most of them directly model the raw student interaction sequence with neural networks, implicitly assuming that all interactions contribute to the evolution of students’ knowledge states in a same manner~\cite{survey2023,survey2024}. 
However, such a unified modeling paradigm overlooks the fact that students may exhibit substantially different learning behaviors at different phases of knowledge acquisition.
A recent study DisKT\cite{diskt} decomposes students' response sequences based on response correctness, and learns separate representations of familiar and unfamiliar abilities from the respective subsequences.
However, this correctness-based decomposition is too coarse to represent the behavioral semantics of different learning phases, such as initial knowledge acquisition versus skill consolidation.

\begin{figure}[t] 
    \centering 
    \includegraphics[width=1\linewidth]{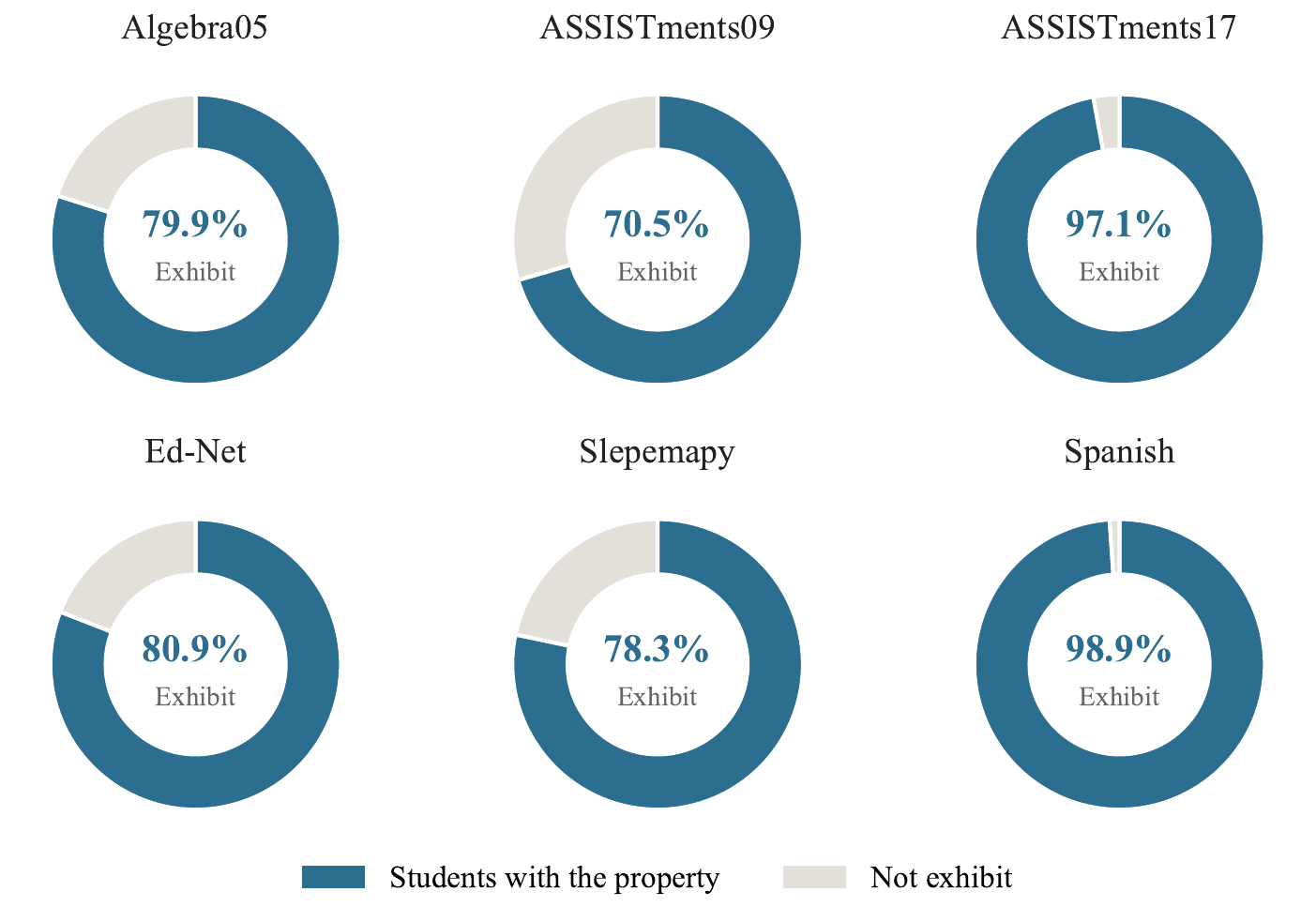} 
    \caption{Proportion of students exhibiting the two-phase learning property across six benchmark datasets. 
    A student is considered to exhibit the property if their late-phase accuracy exceeds early-phase accuracy on more than a quarter of qualifying knowledge concepts.
     } 
    \label{fig:learning_property} 
\end{figure}

In practice, a student’s response behavior before and after mastering a knowledge concept can be quite different.
Early interactions mainly reflect the learner’s ability development through repeated practice, while later interactions better indicate the stability and utilization of mastered knowledge.
As shown in Fig. \ref{fig:learning_property}, a substantial proportion of students are more likely to answer knowledge concepts correctly in later interactions after sufficient practice.
Therefore, directly encoding the entire interaction sequence without distinguishing these phases may mix complex learning signals and limit the model’s ability to precisely characterize the dynamic evolution of learners’ knowledge phases.
This phenomenon raises a challenging question:
\textit{How can we effectively model phase-specific learning behaviors from raw student interaction sequences to better characterize the dynamic evolution of learners’ knowledge states?}

To this end, we propose Phase-Aware Knowledge Tracing (\name), a new KT model that decomposes the student interaction sequence into ability and proficiency phases for phase-specific modeling. 
The learning process comprises two phases. 
In the initial phase, students gradually master knowledge concepts through practice. 
Once mastery is achieved, further practice drives skill automation, leading to more fluent and consistent performance rather than additional gains in correctness. 
Accordingly, our decomposition mechanism partitions each learner's response history based on the cumulative correct response count per knowledge concept, assigning early interactions to the ability phase and later ones to the proficiency phase. 
The two decomposed subsequences, together with the original complete sequence, are fed into a multi-branch Transformer-based backbone, where separate decoders encode the ability, proficiency, and holistic aspects of the student's knowledge state.
A type-aware readout module then integrates these complementary representations into a unified knowledge state for final prediction.
The contributions of this paper are summarized as follows:
\begin{itemize}
    \item We develop a new learning behavior-aware decomposition mechanism that partitions student interaction sequences into ability and proficiency phases. 

    \item We propose a multi-branch Transformer-based backbone with a type-aware readout module to model the decomposed interaction sequences for KT tasks.
    \item Extensive experiments on six public benchmarks demonstrate that \name achieves consistent improvements over all representative baselines.
\end{itemize}

\section{Related Work}
In this section, we first provide a comprehensive review of recent efforts in KT.
Then, we introduce the approaches of applying sequence decomposition mechanism in the KT task.
\subsection{Knowledge Tracing}
Prior approaches for KT tasks are built on probabilistic models and psychometric theory.
BKT\cite{bkt}, a classic probabilistic model, employs a hidden Markov model to simulate a student's knowledge mastery and makes predictions based on their historical exercise performance. 
TLS‑BKT\cite{tlsbkt} applied stage split based on BKT, but it still remains within the probabilistic graphical modeling framework and does not leverage the representational power of deep neural networks.
Item Response Theory (IRT)\cite{irt}, a psychometric model, estimates the probability of a correct response by modeling the interaction between a person's latent ability (e.g., knowledge level) and item characteristics such as difficulty and discrimination. 
However, these methods rely on strong assumptions and manually defined knowledge structures, which restricts their applications in the real-world scenario\cite{irt-review}.

Recently, deep learning-based approaches for KT tasks have attracted great attention due to the superior performance and flexibility. 
DKT\cite{dkt} usher in a new era for KT by introducing RNNs to directly model sequential student interactions.
Then, KT models try to represent knowledge states more explicitly by introducing more complex neural networks. 
DKVMN\cite{dkvmn} is proposed with a memory-augmented network with separate matrices for concepts and mastery states, while AKT\cite{akt} and SAINT\cite{saint} leverage Transformer\cite{transformer} attention to dynamically weigh historical interactions.

Building upon these advancements in sequential modeling, subsequent research explore ways to incorporate external knowledge structures and novel training paradigms to further enhance model capability.
Besides sequential models, advanced techniques in deep learning have been introduced into KT.
GKT\cite{gkt}, a graph-based KT model, operates on a prerequisite graph using Graph Neural Networks (GNNs) to capture dependencies between knowledge concepts.
To enhance model robustness, CL4KT\cite{cl4kt} employs contrastive learning to derive noise-invariant student representations, and DTransformer\cite{dtransformer} integrates diagnostic decoders to isolate concept-specific mastery levels.

To enhance interpretability, educational theories have been integrated into deep KT models.
Deep-IRT\cite{deepirt} integrates the classic Item Response Theory (IRT) into a deep learning framework, allowing the model's parameters (e.g., ability, difficulty) to retain their psychometric meanings.
MIKT\cite{mikt} explicitly models the monotonic relationship between a student's knowledge state and their probability of answering correctly, a core principle in education that ensures interpretable predictions.
CoreKT\cite{corekt} focuses on concept-level interpretability by designing a model that guarantees monotonicity at the level of individual knowledge components, making its predictions more transparent and educationally plausible.

In parallel, a contrasting line of work seeks simplicity and efficiency. 
SparseKT\cite{sparsekt} tackles data sparsity in long sequences via a sparse Transformer architecture, replacing full attention with localized patterns to improve efficiency and generalization.
SimpleKT\cite{simplekt} advocates for model simplicity, showing that a well-regularized, single-layer LSTM can match the performance of far more complex KT models.

In summary, prior KT models largely adopt a homogeneous modeling paradigm, applying a single system to the entire learning sequence. 
This approach inherently conflates the distinct dynamics of early knowledge acquisition (driven by cognitive ability) and later skill consolidation (driven by practice). 
Our work breaks from this tradition by introducing a temporally-grounded, data-driven decomposition of the sequence, enabling explicit and separate modeling of these two fundamental learning phases.
\subsection{Decomposition Mechanism in Knowledge Tracing}
Introducing sequence decomposition into KT tasks is a promising way to capture detailed features in learning process.
DisKT \cite{diskt} pioneers the application of sequence decomposition in KT by employing structural causal model to deconfound spurious correlations and estimate counterfactual effects. 

However, its intervention mechanism relies on a simplistic partitioning of student interactions (e.g., by historical performance) and fails to model the distinct educational semantics, such as the phase of initial knowledge acquisition versus skill consolidation.

In contrast, the proposed \name introduces a tailored decomposition strategy that divides the history records into ability phase and proficiency phase, facilitating the causal inference-based technique deeply modeling the learning behaviors.

\section{Preliminaries}
\subsection{Problem Formulation}
\label{subsec:problem_formulation}

The KT task aims to model the evolution of a student's latent knowledge state based on their observed learning interactions. We begin by defining the core elements of an educational system:

\begin{itemize}
    \item $\mathcal{S} = \{s_1, s_2, \dots\}$: the set of all students.
    \item $\mathcal{Q} = \{q_1, q_2, \dots\}$: the set of all questions.
    \item $\mathcal{C} = \{c_1, c_2, \dots\}$: the set of all knowledge concepts (or skills).
\end{itemize}
For a given student $s \in \mathcal{S}$, their learning process over $T$ discrete time steps is recorded as a sequential interaction history:
\[
X^{(s)} = \{x_1, x_2, \dots, x_T\}.
\]
Each interaction $x_t$ at time step $t$ is represented as a triple:
\[
x_t = (q_t, \mathcal{C}_{q_t}, r_t),
\]
where:
\begin{itemize}
    \item $q_t \in \mathcal{Q}$ denotes the question answered at step $t$.
    \item $\mathcal{C}_{q_t} \subseteq \mathcal{C}$ is the set of knowledge concepts assessed by question $q_t$, i.e., $\mathcal{C}_t = \{c^{(1)}_t, c^{(2)}_t, \dots, c^{p}_t\}$.
    \item $r_t \in \{0, 1\}$ is the binary response label, with $r_t=1$ indicating a correct answer and $r_t=0$ indicating an incorrect one.
\end{itemize}

Formally, the KT task is defined as follows: given a student's historical interaction sequence $X^{(s)}$ up to time $T$, and the next question $q_{T+1}$ (along with its associated concepts $\mathcal{C}_{T+1}$), the goal is to predict the probability that the student will answer $q_{T+1}$ correctly:
\[
P(r_{T+1} = 1 \mid X^{(s)}, q_{T+1}, \mathcal{C}_{q_{T+1}}).
\]
This probability reflects the model's estimation of the student's mastery of concepts in $\mathcal{C}_{T+1}$ at the moment of prediction, conditioned on their entire learning trajectory.

\section{Methodology}
\begin{figure*}[htbp] 
    \centering 
    \includegraphics[width=1\textwidth]{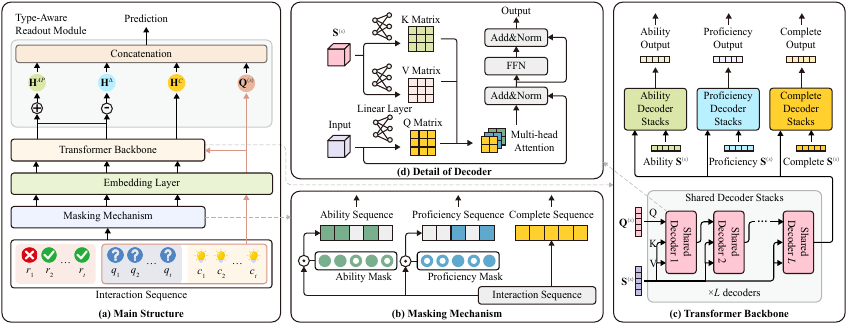} 
    \caption{An overview of the \name model. The response correctness, question ID, and concept ID sequences are processed through the mask mechanism and Embedding Layer into the Transformer-based Backbone. The highlighted path (red box and arrow) in Fig. \ref{fig:model_graph} (a) indicates that the question ID and concept ID are fed solely into the embedding layer, whose output directly serves as the question feature representations $\mathbf{Q}^{(s)}$. The three-branch outputs of the Backbone, $\mathbf{H}^{A}$, $\mathbf{H}^{P}$, and $\mathbf{H}^{C}$, together with $\mathbf{Q}^{(s)}$, are subsequently aggregated by the Type-Aware Readout Module to produce the final prediction.} 
    \label{fig:model_graph} 
\end{figure*}


In this section, we detail our proposed \name, which includes three core components: learning behavior-aware sequence decomposition, multi-branch Transformer-based backbone and type-aware readout module.
These components are utilized for generating input sequences, learning representation and producing the final representations for prediction, respectively.
After introducing the above components, we provide the theoretical analysis of \name from the perspective of causal inference.

\subsection{Learning Behavior-Aware Sequence Decomposition}

During the initial learning of a knowledge concept (KC), students first aim to master the core concept and practice it repeatedly to master its application across different contexts.
The repeated practice increases their familiarity, thereby raising the probability of a correct response involving the specific KC.
Subsequently, further practice leads to skill automation \cite{human_performance}, as students have internalized the common solution patterns for this KC.
This illustrates that proficiency is dynamically built up over time.
In summary, the dominant factor influencing a student's performance undergoes a shift: from initial knowledge acquisition to the consolidation and automation of skills through practice.
This situation motivates us to decompose the raw interaction sequence into separate subsequences containing different learning phases to deeply understand the learning behaviors of the target student.

Specifically,  we decompose the learning process into two phases, the ability phase and the proficiency phase:

\begin{itemize}
    \item \textbf{Ability Phase: }This initial phase captures the period when a student encounters a new knowledge concept and is actively working to understand and master its core principles. 
    During this phase, the student's performance is primarily governed by their intrinsic cognitive aptitude for grasping the concept.
    
    \item \textbf{Proficiency Phase: }At this stage, the core knowledge is assumed to be acquired, and performance is increasingly driven by the effects of practice, leading to skill automation and fluency. 
    The student's probability of a correct answer becomes more influenced by the recency, frequency, and variety of practice rather than initial aptitude. 
    This phase is therefore optimal for modeling the dynamic, practice-induced factor of proficiency, which captures how well a student can reliably and efficiently apply the knowledge across different problem contexts.
\end{itemize}

Based on the descriptions of the above two phases in the learning process, we now detail the generation mechanism of interaction sequences for each phase.
Given the initial interaction sequence of the student $s$, we first transform the raw response sequence into the KC-based interaction sequences according to the previous study~\cite{simplekt, diskt}.
Then, for each KC $c$ at the $t$-th step in the KC sequence,
we define a binary \textit{phase indicator function} $\delta(s, c, t)$ based on a decomposing threshold $k$ and the cumulative correct attempts:
\begin{equation}
    \delta(s, c, t) = \mathbb{I}\big(\gamma(s, c, t) < k\big),
\end{equation}
where $\gamma(s,c,t)$ denotes the number of times student $s$ has correctly answered questions related to concept $c$ at the $t$-th step, and $k$ is a hyper-parameter.
$\mathbb{I}(\cdot) \in \{0, 1\}$ denotes the indicator function.

According to the output of the phase indicator function $\delta(s, c, t)$, we further develop two decomposition mask matrices $\mathbf{M}^{(s),A}\in \mathbb{R}^{L^{KC}_{s}}$ and $\mathbf{M}^{(s),F}\in \mathbb{R}^{L^{KC}_{s}}$ to achieve the learning behavior-aware sequence decomposition, where $L^{KC}_{s}$ denotes the length of the student's KC sequence, $\mathbf{M}^{(s),A}$ and $\mathbf{M}^{(s),F}$ are adopted for the ability phase and the proficiency phase, respectively.
Specifically, for the given interaction KC tuple $(c,t)$, the corresponding values in $\mathbf{M}^{(s),A}$ and $\mathbf{M}^{(s),F}$ are defined as follows:
\begin{equation}
    \mathbf{M}^{(s),A}_t = \delta(s, c, t), \quad \mathbf{M}^{(s),F}_t = 1 - \delta(s, c, t).
    \label{eq-mask}
\end{equation}

\begin{figure}[t] 
    \centering 
    \includegraphics[width=1\linewidth]{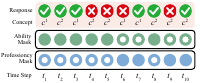} 
    \caption{
    The illustration of the decomposition mask generation.
    Taking the cumulative correct answer threshold \(k=2\) as the dividing criterion: for concept \(c^1\) and \(c^2\), the cumulative correct responses first reach twice at time step \(t_3\) and  \(t_8\), respectively. Thus the interactions at \(t_1,t_3\) and \(t_8\) are divided into the ability phase, while the interactions at \(t_6,t_7,t_9\) and \(t_{10}\) belong to the proficiency phase. } 
    \label{fig:mask} 
\end{figure}

Based on the Eq. (\ref{eq-mask}), an interaction at time $t$ involving concept $c$ is assigned to the proficiency phase if and only if the cumulative correct count for that concept has reached or exceeded the threshold $k$. 
Otherwise, this KC tuple will be assigned to the ability phase.
This binary decomposing ensures that each interaction for a given concept exclusively belongs to one of the two phases, thereby enabling separate modeling of the ability and proficiency components throughout the learning sequence.
For simplicity, we consider only the scenario where each item contains one knowledge concept. Having multiple knowledge concepts does not impact the method's implementation.
Fig.\ref{fig:mask} illustrates the generation of two masks.

\subsection{Multi-Branch Transformer-Based Backbone}
After constructing the decomposition mask matrices, \name further develops a multi-branch Transformer-based backbone to learn the knowledge states of student $s$ from the input interaction sequences via the generated decomposition masks $\mathbf{M}^{(s),A}$ and $\mathbf{M}^{(s),F}$.

Based on previous approaches\cite{akt}\cite{dtransformer}\cite{diskt}, we first construct the initial input features of the KC-based interaction sequences for the student $s$.
Specifically, the initial representations of $t$-th interaction $(c_t, r_t)$ are calculated as follows:
\begin{equation}\label{embedding}
\mathbf{Q}^{(s)}_t=\mathbf{c}_{c_t}+\mathbf{d}_{q_t},\quad \mathbf{S}^{(s)}_t=\mathbf{c}_{c_t}+\mathbf{r}_{r_t}+\mathbf{d}_{q_t},
\end{equation}
where $\mathbf{c}_{c_t}\in \mathbb{R}^d$ is the embedding of concept $c_t$, $\mathbf{d}_{q_t}\in \mathbb{R}^d$ is a learnable vector representing the difficulty of question $q_t$ and $c_t \in \mathcal{C}_{q_t}$.
$\mathbf{r}_{r_t}\in \mathbb{R}^d$ is the embedding of response $r_t$.
$d$ denotes the dimension of these embeddings.
$\mathbf{Q}^{(s)}\in \mathbb{R}^{{L^{KC}_{s}}\times d}$ and $\mathbf{S}^{(s)}\in \mathbb{R}^{{L^{KC}_{s}}\times d}$ are the initial representations of the corresponding KC $c_t$ and the interaction tuple $(c_t, r_t)$.

Then, \name adopts the generated decomposition masks to create three types of input branches for comprehensively capturing the semantic information from the student's learning behaviors:
\begin{equation}
    \mathbf{Q}^{A} = \mathbf{Q}^{(s)}\odot \mathbf{M}^{(s),A},\mathbf{Q}^{F} = \mathbf{Q}^{(s)}\odot \mathbf{M}^{(s),F},\mathbf{Q}^C=\mathbf{Q}^{(s)},
\end{equation}
\begin{equation}
    \mathbf{S}^{A} = \mathbf{S}^{(s)}\odot \mathbf{M}^{(s),A},\mathbf{S}^{F} = \mathbf{S}^{(s)}\odot \mathbf{M}^{(s),F},\mathbf{S}^C=\mathbf{S}^{(s)}.
\end{equation}

The descriptions of the above three branches are as follows:

\begin{itemize}
    \item 
     $\mathbf{Q}^{A}$ and $\mathbf{S}^{A}$ are the ability branch input which processes the subset of interactions identified as belonging to the \textit{ability phase} by the ability mask $\mathbf{M}^{(s),A}$. 
    This sequence primarily reflects the student's intrinsic cognitive capacity for initial knowledge acquisition.
    The corresponding ability representation $\mathbf{H}^A$ is learned from this branch.
    \item 
    $\mathbf{Q}^{F}$ and $\mathbf{S}^{F}$ are the proficiency branch input which operates on the complementary subset marked by the proficiency mask $\mathbf{M}^{(s),F}$, corresponding to the \textit{proficiency phase}. 
    This sequence captures the dynamics of practice consolidation, skill automation, and performance consistency after initial mastery.
    The proficiency representation $\mathbf{H}^F$ is derived from this branch.
    \item 
    $\mathbf{Q}^{C}$ and $\mathbf{S}^{C}$ are the complete branch which processes the original, non-decomposed interaction sequence. 
    It serves to maintain a holistic view of the student's overall learning trajectory. 
    The comprehensive student representation $\mathbf{H}^C$ is obtained from this branch.
\end{itemize}

In practice, we also adopt the response-aware strategy introduced by \cite{diskt} to strengthen the interaction decomposition.
\name further leverages a two-stage Transformer-based backbone to learn the dynamic knowledge states from the input three branches.
Specifically, in the first stage, all input branches share the same Transformer decoder. 
While, in the second stage, each branch is fed to independent Transformer decoder.
The shared first stage learns universal sequential features for efficiency, and dedicated second-stage decoders enable phase-specific specialization for representation learning.

The forward pass of the first decoder layer in this shared stage is defined as:
\begin{equation}
\mathbf{H}^{b}_{1}=\text{Decoder}(Q=\mathbf{Q}^{b},K=\mathbf{S}^{b},V=\mathbf{S}^{b}),
\end{equation}
where $b\in \{A,F,C\}$ represents the branch indicator. 
$\text{Decoder}(\cdot)$ represents the standard Transformer decoder which consists of $L$ Transformer layers.

Similarly, the forward pass of the second decoder layer in phase-specific stage is defined as:
\begin{equation}
\mathbf{H}^{b}=\text{Decoder}(Q=\mathbf{H}^{b}_{1},K=\mathbf{S}^{b},V=\mathbf{S}^{b}),
\end{equation} 
where $\text{decode}^{b}(\cdot)$ represents the standard Transformer decode which consists of $L$ Transformer layers for branch $b$.

\subsection{Type-Aware Readout Module}
After obtaining the three student representations (ability $\mathbf{H}^A$, proficiency $\mathbf{H}^F$, and complete $\mathbf{H}^C$) from the Transformer-based backbone, we design a fusion mechanism to integrate them into a comprehensive student representation.

First, the ability and proficiency representations are adaptively fused via a learnable coefficient $\alpha \in [0,1]$:
\begin{equation}
\mathbf{H}^{AF} = \alpha \cdot \mathbf{H}^A + (1 - \alpha) \cdot \mathbf{H}^F,
\end{equation}
where $\alpha$ is optimized during training to automatically balance the contribution of ability and proficiency.

To capture the differential gain from practice, especially for students who improve significantly after the initial learning phase, we compute the proficiency-ability gap:
\begin{equation}
\mathbf{H}^{\Delta} = \mathbf{H}^F - \mathbf{H}^A,
\end{equation}
This representation encodes the extent to which practice has enhanced performance beyond the student's initial ability.

The final comprehensive student representation $\mathbf{H}^S$ is then constructed by concatenating the fused representation $\mathbf{H}^{AF}$, the complete-branch representation $\mathbf{H}^C$, and the differential gain vector $\mathbf{H}^{\Delta}$:
\begin{equation}
    \mathbf{H}^S = [\mathbf{H}^{AF} \| \mathbf{H}^C \| \mathbf{H}^{\Delta}].
\end{equation}
where $[\cdot \| \cdot]$ denotes vector concatenation. This design ensures the model retains information from the holistic learning trajectory while preserves the semantic behavioral features from different learning phases.

\subsection{Response Prediction and Model Training}
The prediction layer takes the concatenated joint representation $\mathbf{z} = [\mathbf{H}^S \| \mathbf{Q^{(s)}}] \in \mathbb{R}^{4d}$ which combines the representations of the student and the target question as input and outputs the predicted probability of a correct response. 
This layer is implemented as a multi-layer perceptron (MLP) with non-linear activation:
\begin{equation}
     \hat{y} = \sigma(\text{MLP}(\mathbf{z})),
\end{equation}
where $\sigma(\cdot)$ denotes the sigmoid activation function, $\hat{y} \in (0,1)$ denotes the predicted probability that the student will answer the question correctly.

The model is trained using a composite loss function that combines two objectives.
The primary component is the binary cross-entropy loss between predicted and actual response outcomes:
\begin{equation}
\mathcal{L}_{\text{BCE}} = -\frac{1}{N} \sum_{i=1}^{N} \left[ y_i \log(\hat{y}_i) + (1 - y_i) \log(1 - \hat{y}_i) \right],
\end{equation}
where $y_i \in \{0,1\}$ represents the ground truth correctness label, $\hat{y}_i \in [0,1]$ is the model's predicted probability, and $N$ is the number of valid training samples.

To accelerate the model convergence and explicitly guide the ability branch to learn distinct skill learning patterns, we introduce a regularization term applied specifically to this branch:
\begin{equation}
\mathcal{L}_{\text{reg}} = -\frac{1}{U} \sum_{u=1}^{U} \left\| \mathbf{S}_u^{A+} - \mathbf{S}_u^{A-} \right\|_2,
\end{equation}
where $\mathbf{S}_u^{A+}$ and $\mathbf{S}_u^{A-}$ represent the processed embeddings for positive (correct) and negative (incorrect) samples for ability branch within batch $u$, respectively. 

The final loss function combines the binary cross-entropy loss with the regularization term:
\begin{equation}
\mathcal{L}_{\text{total}} = \mathcal{L}_{\text{BCE}} + \mathcal{L}_{\text{reg}}.
\end{equation}

\subsection{Causality Analysis on \name}
\begin{figure}[htbp] 
    \centering 
    \includegraphics[width=1\linewidth]{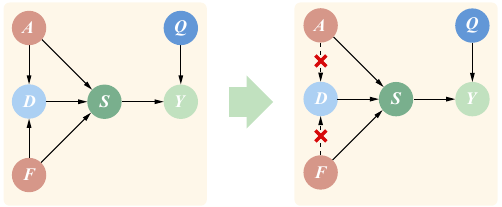} 
    \caption{The causal graph of conventional KT model and DAPKT, where the direct causal paths from ability ($A$) and proficiency ($F$) to student features ($S$) are explicitly cut.} 
    \label{fig:causal_graph} 
\end{figure}

To deeply understand the proposed \name, we provide the causality analysis~\cite{dice, decrs} on \name.
Specifically, we construct a structural causal model that identifies the factors influencing student performance.
As shown in Fig. \ref{fig:causal_graph}, the colliding effect between a student's intrinsic ability and acquired proficiency on student features introduces confounding bias, a key limitation conflating these factors in traditional KT models.

\subsubsection{Causal Graph}
In the structural causal graph, we define the following key variables and their causal relationships:
\begin{itemize}
    \item$A$ denotes the student's cognitive ability.

    \item$F$ denotes the student's proficiency on knowledge concepts.

    \item$D$ denotes the distribution of the student's historical interaction sequence.

    \item$S$ denotes latent vector that encodes the student's features.

    \item$Y$ denotes the predicted probability of the student answering a given question correctly.

    \item$Q$ denotes feature vector representing the characteristics of a question.
    
    \item$(A, F)\rightarrow D$ refers to A and F jointly shape their historical response pattern D. Students with higher ability tend to provide correct answers more rapidly during the initial learning phase, while greater proficiency enables more consistent correct performance after mastery, collectively determining the observed distribution of interactions.

    \item$(A, F, D)\rightarrow S$ refers to modeling the process in conventional KT approaches, where the student representation $S$ is learned by agglomerating the entire interaction history D, thereby entangling the distinct influences of intrinsic ability A, acquired proficiency F, and other student-specific traits. 

    \item$(S, Q)\rightarrow Y$ denotes that the predicted response probability $Y$ is determined by the joint effect of the student's features $S$ and the question characteristics $Q$.
\end{itemize}

Crucially, ability and proficiency exert both direct and indirect effects on the student features:
\begin{itemize}
    \item Effects: Ability and proficiency can directly impact the student features, independent of the current representation S (e.g., through factors like test-taking confidence or quick recall).

    \item Effects (Mediated by S):More importantly, ability and proficiency shape the student representation S over time. This evolving representation S, in turn, directly influences the prediction Y. Therefore, the total effect of A and F on Y is composed of both their direct effects and their indirect effects mediated through S, accounting for their dual contributions to the outcome.
\end{itemize}

In summary, a student's performance is co-determined by their learning ability and practice-induced proficiency, which exhibit both complementarity (e.g., proficiency can compensate for lower ability) and transferability across related knowledge concepts or problem formats.
However, conventional knowledge tracing models conflate these two distinct causal factors into a single, entangled representation of student ability. 
This conflation introduces confounding bias that hinders accurate prediction. 
By explicitly disentangling and separately modeling these two factors, we enable the model to isolate their distinct causal effects, thereby enhancing predictive accuracy.

Next, we will elucidate why distinguishing between learning ability and proficiency can lead to superior predictive performance from the perspective of probability modeling.
\begin{subequations}
    \begin{align}
        &P(Y|S=s, Q=q)\label{prob_eq1} \\
        = &\sum_{a\in \mathcal{A}}\sum_{f\in \mathcal{F}}\sum_{d\in \mathcal{D}}\frac{P(Y,s,q,d,a,f)}{P(s,q)}\label{prob_eq2} \\
        = &\sum_{a\in \mathcal{A}}\sum_{f\in \mathcal{F}}\sum_{d\in \mathcal{D}}\frac{P(a)P(f)P(d|a,f)P(s|a,f,d)P(q)P(Y|s,q)}{P(s)P(q)}\label{prob_eq3}\\
        = &\sum_{a\in \mathcal{A}}\sum_{f\in \mathcal{F}}\sum_{d\in \mathcal{D}}P(a,f|d)P(d)P(Y|S(a,f,d),q)\label{prob_eq4}\\
        =&P(a_s,f_s)P(Y|S(a_s,f_s,d_s),q)\label{prob_eq5}
    \end{align}
\end{subequations}
Eq.(\ref{prob_eq2}) and Eq.(\ref{prob_eq3}) are derived by applying the law of total probability, the structure of the causal graph, and Bayes' rule. Given that S is a deterministic function of A, F, and D, we obtain Eq.(\ref{prob_eq4}). By focusing our analysis on the subspace $d_s$ conditioned on specific ability $a_s$ and proficiency $f_s$ values, we arrive at the final estimable form in Eq.(\ref{prob_eq5}), where $\mathcal{D}$, $\mathcal{A}$, $\mathcal{F}$ denote the sample spaces of D, A, P respectively.

The probabilistic derivation in Eq.(\ref{prob_eq5}) demonstrates that the student feature $S$ is confounded by the backdoor path $A, F \rightarrow D \rightarrow S$. 
This means the effects of intrinsic ability ($A$) and acquired proficiency ($F$) are entangled through the observed historical interaction pattern $D$. 
This dual-source confounding creates ambiguity in predictions: the model cannot determine if a high prediction stems from the student's ability, proficiency, or a mixture of both.
Consequently, this entanglement introduces spurious correlations and adversely impacts prediction accuracy.

To address this, explicitly disentangling ability and proficiency allows the model to isolate their distinct causal effects. 
This separation breaks the spurious correlation, enabling a more accurate estimation of each factor's true contribution to boost the performance. 

\section{Experiments}
\subsection{Baselines} 
In this paper, we adopt the following 6 representative methods for KT as the baselines:
\begin{itemize}
    \item \textbf{DKT}\cite{dkt}: DKT is the first model that employs RNNs to establish a foundational deep learning baseline.
    \item \textbf{DKVMN}\cite{dkvmn}: DKVMN introduces a memory-augmented neural network that uses a static key matrix and a dynamic value matrix to explicitly track and update students' evolving knowledge states.
    \item \textbf{SAKT}\cite{sakt}: SAKT employs a self-attention mechanism within a Transformer architecture to dynamically weigh past learning interactions. 
    \item \textbf{SparseKT}\cite{sparsekt}: SparseKT addresses the data sparsity challenge in KT by explicitly leveraging the sparsity of student interaction sequences. 
    \item \textbf{SimpleKT}\cite{simplekt}: SimpleKT advocates for architectural simplicity and efficiency in KT by revisiting and simplifying the foundational DKT framework.
    \item \textbf{DisKT}\cite{diskt}: DisKT pioneers the integration of causal inference into knowledge tracing to deconfound spurious correlations in learning data. 
\end{itemize}

\subsection{Datasets}
Our experiments are conducted on 6 widely-used real-world datasets: Algebra05, Assistments09, Assistments17, Ed-Net, Slepemapy, and Spanish. The statistical information of the datasets after pre-processing is shown in Table \ref{tab:datasets}.
\begin{table}
\centering
\caption{Statistics of 6 datasets.}
\label{tab:datasets}
\begin{tblr}{
  cells = {c},
  hline{1,8} = {-}{2pt},
  hline{2} = {-}{1pt},
}
Dataset   & \#students & \#questions & \#concepts & \#interactions \\
Assist09  & 3,644      & 17,727      & 123        & 281,890        \\
Assist17  & 1,079      & 3,162       & 102        & 942,816        \\
Ed-Net    & 5,000      & 12,117      & 189        & 676,276        \\
Spanish   & 182        & 409         & 221        & 578,726        \\
Slepemapy & 5,000      & 2,723       & 1,391      & 625,523        \\
Algebra05 & 572        & 173,650     & 112        & 609,971        
\end{tblr}
\end{table}
\begin{table*}[t!]
\centering 
\caption{\textbf{Performance comparison on all datasets. The reported values represent the mean from 5-fold cross-validation. The best performance is highlighted in bold, and the second-best is indicated with underlining. The relative improvement percentages achieved by \name are shown in the bottom row.}}
\label{tab:comparison}
\begin{tblr}{
  cells = {c},
  cell{2}{1} = {r=2}{},
  cell{2}{9} = {font=\bfseries},
  cell{3}{9} = {font=\bfseries},
  cell{4}{1} = {r=2}{},
  cell{4}{9} = {font=\bfseries},
  cell{5}{9} = {font=\bfseries},
  cell{6}{1} = {r=2}{},
  cell{6}{9} = {font=\bfseries},
  cell{7}{9} = {font=\bfseries},
  cell{8}{1} = {r=2}{},
  cell{8}{9} = {font=\bfseries},
  cell{9}{9} = {font=\bfseries},
  cell{10}{1} = {r=2}{},
  cell{10}{9} = {font=\bfseries},
  cell{11}{9} = {font=\bfseries},
  cell{12}{1} = {r=2}{},
  cell{12}{9} = {font=\bfseries},
  cell{13}{9} = {font=\bfseries},
  vline{2-3,9} = {1-14}{0.5pt},
  hline{1,14} = {-}{2pt},
  hline{2} = {-}{1pt},
  hline{4,6,8,10,12} = {-}{0.5pt}
}
Dataset   & Metric & DKT    & DKVMN          & SAKT   & SparseKT       & SimpleKT       & DisKT          & \name  & \%Improv. \\
Algebra05 & AUC    & 0.7751 & 0.7824         & 0.7513 & 0.7882         & 0.7921         & \uline{0.7969} & 0.7987 & 0.22\%    \\
          & ACC    & 0.7961 & 0.7960         & 0.7834 & 0.7955         & 0.7953         & \uline{0.7995} & 0.8017 & 0.27\%    \\
ASSISTments09  & AUC    & 0.7590 & 0.7580         & 0.7305 & 0.7679         & 0.7730         & \uline{0.7775} & 0.7812 & 0.49\%    \\
          & ACC    & 0.7235 & \uline{0.7254} & 0.7054 & 0.7210         & 0.7214         & 0.7220         & 0.7341 & 1.19\%    \\
ASSISTments17 & AUC    & 0.6844 & 0.6781         & 0.6416 & 0.7126         & 0.7125         & \uline{0.7293} & 0.7369 & 1.04\%    \\
          & ACC    & 0.6534 & 0.6533         & 0.6324 & 0.6747         & 0.6738         & \uline{0.6830} & 0.6872 & 0.62\%    \\
Ed-Net     & AUC    & 0.6574 & 0.6611         & 0.6470 & 0.7033         & 0.7062         & \uline{0.7140} & 0.7199 & 0.84\%    \\
          & ACC    & 0.6402 & 0.6410         & 0.6301 & 0.6692         & 0.6661         & \uline{0.6714} & 0.6754 & 0.60\%    \\
Slepemapy & AUC    & 0.7129 & 0.7130         & 0.6716 & 0.7261         & \uline{0.7306} & 0.7299         & 0.7377 & 0.98\%    \\
          & ACC    & 0.7883 & 0.7862         & 0.7802 & \uline{0.7916} & 0.7910         & 0.7906         & 0.7952 & 0.46\%    \\
Spanish  & AUC    & 0.8298 & 0.8250         & 0.8047 & 0.8366         & 0.8430         & \uline{0.8453} & 0.8566 & 1.33\%    \\
          & ACC    & 0.7774 & 0.7800         & 0.7609 & 0.7826         & 0.7893         & \uline{0.7904} & 0.7998 & 1.19\%    
\end{tblr}
\end{table*}
\begin{itemize}
    \item \textbf{Algebra05}\cite{algebra05}: The Algebra05 dataset originates from the KDD Cup 2010 EDM Challenge and contains detailed step-level student responses to algebra questions.
    \item \textbf{ASSISTments09}\cite{assist}: The ASSISTments09 dataset is a foundational benchmark collected from the ASSISTment intelligent tutoring system during the 2009-2010 school year. 
    It primarily consists of mathematics exercises and has been extensively utilized to evaluate the performance of knowledge tracing models. 
    \item \textbf{ASSISTments17}\cite{assist}: The ASSISTments17 dataset has diverse problem types and inherent noise present a significant challenge for modeling complex, long-sequence learning.
    \item \textbf{Ed-Net}\cite{ednet}: The Ed-Net dataset from the Santa tutoring platform, is the largest public educational interaction dataset, containing fine-grained logs of millions of student problem-solving attempts. 
    \item \textbf{Slepemapy}\cite{slepemapy}: The Slepemapy dataset from an adaptive geography practice platform, focuses on the factual recall of country locations, capitals, and flags.
    \item \textbf{Spanish}\cite{spanish}: The Spanish dataset records middle-school students' practice on vocabulary and grammar over a 15-week semester.
\end{itemize}
\subsection{Implementation Details}
We employ a comprehensive experimental setup to ensure robust evaluation. 
All models are evaluated using 5-fold cross-validation, where the data in each fold is decomposed into 80\% for training, 10\% for validation, and 10\% for testing. 
The Adam optimizer is used for training all models. 
To mitigate overfitting, an early stopping strategy is applied, monitoring the validation performance with a patience of 10 epochs. 
To guarantee a fair comparison, we conduct a hyperparameter search for each model on every dataset. 
Key parameters, including the learning rate, dropout\cite{dropout} rate, embedding dimension, and 
the hidden dimension, are tuned within a limited, predefined search space to identify the optimal configuration for each model-dataset pair based on validation set performance.
All experiments are conducted on a Linux server using 2 NVIDIA GeForce RTX 3090 GPUs and all models are implemented in PyTorch. Our code is available at https://github.com/ThaliaBee/PAKT.

\subsection{Performance Comparison}
Following previous studies~\cite{diskt,sparsekt}, we adopt two  metrics as our evaluation metrics: Area Under the ROC Curve (AUC) and Accuracy (ACC).
The overall performances of all models, are summarized in Table \ref{tab:comparison}. Several observations can be made from the result as below.

As shown in Table \ref{tab:comparison}, \name achieves the best performance across all six datasets on both AUC and ACC metrics.
The performance advantage spans diverse educational contexts, including mathematics (Algebra05, ASSISTments09, ASSISTments17), language learning (Spanish, Ed-Net), geography (Slepemapy).
These improvements demonstrate the effectiveness of stage-based modeling principle for the knowledge tracing task.
It is worth noting that DisKT~\cite{diskt}, which also decomposes the input sequence, generally achieves competitive results compared to conventional KT models.
However, this improvement is not stable: on certain datasets, DisKT underperforms relative to methods that do not employ sequence decomposing (e.g., SimpleKT and SparseKT).
This phenomenon stems from the fact that a simple correctness-based split only provides a coarse separation of the interaction data, without capturing the temporal characteristics in learning process.

In contrast, \name introduces a fine-grained decomposing strategy.
It explicitly models the temporal transition from ability phase to proficiency phase for each knowledge concept.
Furthermore, the multi-branch backbone and the type-aware readout module fully leverage the complementary information from both phases.
As a result, \name delivers stable and universal improvements over all baselines across every dataset.
The AUC gains over the strongest baseline range from 0.22\% to 1.33\%.

The different magnitudes of improvement can be explained by different characteristics of datasets.
On the one hand, \name achieves more stable and significant gains on datasets with more interactions and a moderate number of knowledge concepts.
\name achieve +1.33\% AUC on Spanish, +1.04\% AUC on ASSISTments17, respectively.
Spanish is characterized by moderate conceptual density and high conceptual interaction coverage, with students interacting with 213.8 out of 221 concepts on average.
ASSISTments17 has the largest amount of interactions.
In such settings, the phase shift from ability to proficiency is more clearly manifested in the data, allowing our stage-based decomposition to fully exploit its modeling capacity.

On the other hand, the smallest improvement is observed on Algebra05 (+0.22\% AUC).
We attribute this limited gain to the low concept coverage within each student's interaction.
A statistical analysis reveals that each Algebra05 student interacts with only 41.3 of its 112 concepts on average on the whole dataset.
When very few concepts enter the proficiency phase, the proficiency branch receives sparse training signal.
More than that, Algebra05 contains 173,650 questions, nearly ten times of ASSISTments09.
This results in extremely sparse training signals per question, leaving the parameters related to question difficulty $\mathbf{d}_{q_t}$ in Eq.(\ref{embedding}) severely undertrained and thus impairing its representational capacity.

\subsection{Sensitivity Analysis of the Threshold Parameter $k$}
\label{k_study}

\begin{figure*}[t] 
    \centering 
    \includegraphics[width=1 \textwidth]{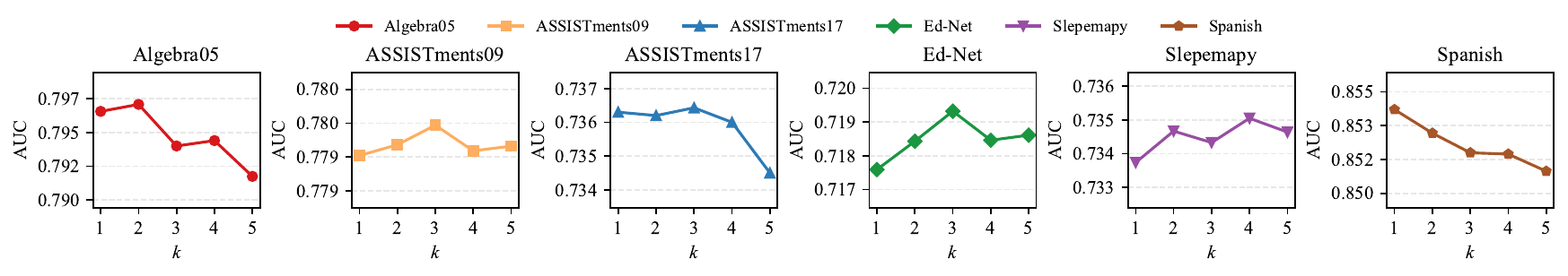} 
    \caption{Effect of parameter $k$ on model performance .} 
    \label{fig:k_study} 
\end{figure*}

Since the threshold $k$ is a core hyper-parameter of our decomposing mechanism, we conducted experiments on the value of $k$.
$k$ determines the dividing point between the ability and proficiency phases.
Specifically, a larger $k$ delays the transition point.
We conduct a sensitivity analysis to understand the impact of $k$ and to identify its optimal operating range.
In our experimental setup, sequences are padded or truncated to a fixed length of 200, consistent with the main experiments.
The results on all six datasets are visualized in Figure~\ref{fig:k_study}.
Our findings are below:

First, across all datasets, performance peaks within a small range ($k \leq 5$).
Setting $k$ too high forces most interactions to remain in the ability phase, causing proficiency phase signal sparser.

Second, the optimal $k$ varies across datasets.
On Spanish and Algebra05, performance is highest at a small $k$ and generally declines thereafter.
Spanish ($k=1$) exhibits a strictly monotonic decrease.
In this domain with 221 concepts, slightly more than most datasets mentioned above, a student's first correct attempt is already a strong signal of mastery. 
Raising $k$ further only dilutes the sparse proficiency interactions with noise.
Algebra05 exhibits a similar downward trend, although with a slight peak at $k=2$.
Due to its significantly fewer proficiency-active concepts, increasing $k$ more severely reduces the number of interactions in the proficiency phase.

On ASSISTments09, ASSISTments17, and Ed-Net, AUC follows a clear rise-then-fall pattern.
Very low $k$ ($k=1,2$) leads to premature decomposing, while $k=3$ provides sufficient evidence of stable mastery without over-delaying the transition.

Slepemapy nominally peaks at $k=4$, but the AUC variation across all $k$ values is the smallest among all datasets (range $< 0.002$), indicating that the model is largely insensitive to the threshold choice.
The platform's large concept pool, that is 1,391 concepts, severely dilutes the per-concept learning signal. 
Compounding this, its high average correct rate (80.9\%) indicates that most presented concepts are already within a student's competency range, which leaves little behavioral contrast between learning phases regardless of the threshold $k$.

\subsection{Performance Across Different Sequence Length Groups}
\begin{figure*}[t] 
    \centering 
    \includegraphics[width=1 \textwidth]{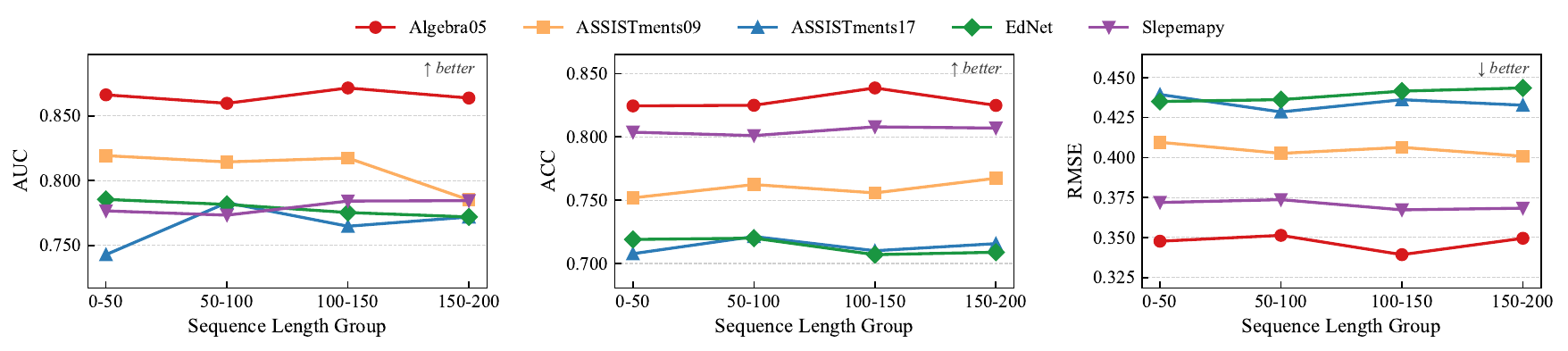} 
    \caption{Performance of \name Across Different Sequence Length Groups.} 
    \label{fig:length_study} 
\end{figure*}
The effectiveness of \name is closely tied to sequence length, since it directly shapes how fully the ability-to-proficiency transition is observable within the modeled window.

The Spanish dataset is excluded, as all student sequences exceed the maximum input length and are truncated, placing every sample in the 150–200 group.
Across nearly all datasets, the variation in AUC across length groups remains within approximately 0.03, indicating that the model maintains robust predictive performance.
Within this overall stability, sequences in the 50–100 and 100–150 groups tend to yield the best performance, as they are long enough to reveal meaningful stage-based progression while remaining untruncated.

Sequences in the 0–50 group lack sufficient interactions to capture the complete ability-to-proficiency developmental trajectory.
For sequences in the 150–200 group, the model truncates from the front to fit the maximum input length, discarding the student's earliest learning interactions.
This means the model loses access to early concept-acquisition behaviors, impairing its ability to characterize students' initial ability-phase interactions.
These findings confirm that a longer raw interaction history does not translate to better performance.

\subsection{Ablation Study}

\begin{figure*}[t] 
    \centering 
    \includegraphics[width=1\textwidth]{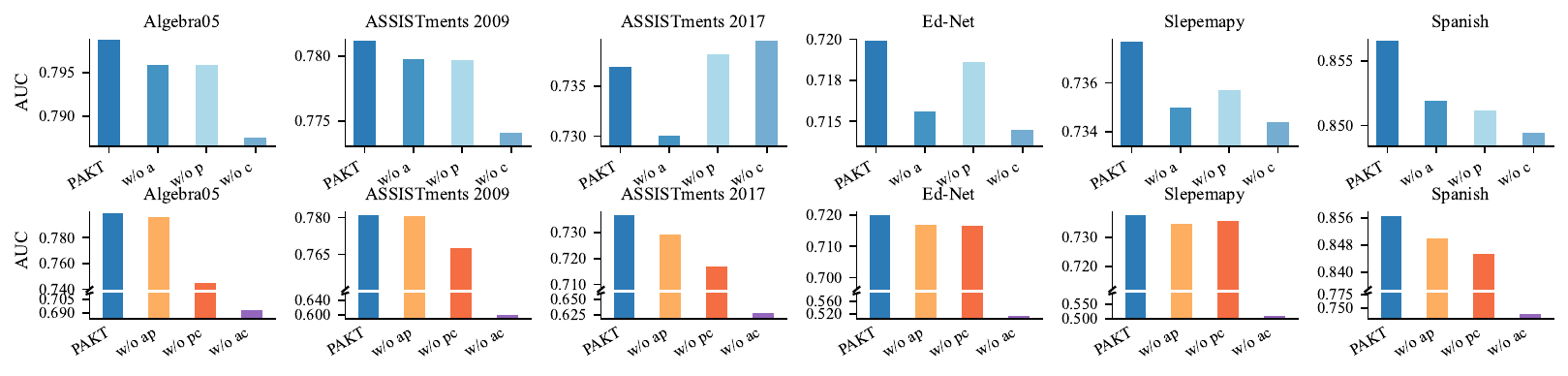} 
    \caption{Ablation study on 6 datasets.} 
    \label{fig:ablation} 
\end{figure*}

To systematically investigate the contribution of each component in \name, we construct seven variants by selectively removing one or two branches from the full model, as shown in Fig \ref{fig:ablation}.

As for the individual component contributions, generally speaking, removing the complete branch causes the largest average degradation among all single-branch ablations, from -0.45\% to -1.38\%, confirming it as the most indispensable component.
The holistic interaction sequence preserves the fundamental information in students' interaction with the most abundant details.
The phase-based branches cannot fully recover these details by reconstruction.
Then, removing the ability branch or the proficiency branch also degrade the model performance on all datasets, which indicates the contributions of these branches for the KT task.

Moreover, as for the two-branch ablations, we can have the similar observations.
For instance, retaining only the Complete branch (w/o ap) can achieve competitive performance on all datasets.
Furthermore, it can be observed that retaining the Ability branch (w/o pc) yields better overall performance than retaining the Proficiency branch (w/o ac). This indicates that the ability branch is more important for characterizing students' knowledge states.

Overall, these results validate the three-branch design: the Complete branch provides the foundational signal, while the Ability and Proficiency branches deliver complementary phase-aware refinements that neither can substitute for the other.

\section{Conclusion}
In this paper, we propose \name, a KT model that incorporates two-phase learning strategy into the KT task to model the dynamics of knowledge acquisition and proficiency consolidation.
\name decomposes each student's interaction sequence into an ability phase and a proficiency phase, reflecting the behavioral semantics of distinct learning stages.
A multi-branch Transformer backbone with a two-stage parameter-sharing mechanism then learns separate knowledge state representations from the ability, proficiency, and holistic branches.
Student representations from different aspects are fused via a readout module for prediction.
\name achieves consistent improvements in both AUC and ACC over all baselines across six public benchmarks. 

However, \name has several limitations.
First, when a student practices too few interactions with a knowledge concept, the cumulative correct response count does not reach the threshold $k$, leaving little or no proficiency-phase signal for that concept and thus weakening the proficiency branch's effectiveness.
Second, the optimal value of $k$ must be manually tuned and varies across datasets.
The observed AUC gains underscore the value of phase decomposition and motivate future investigation into automatic decomposition strategies as well as more robust handling of short interaction sequences.


\section*{Acknowledgments}
This work is supported by the National Natural Science Foundation of China (No. 62507035), the Wuhan University Undergraduate Education Quality Construction Comprehensive Reform Project (Future Curriculum), and the Teaching Reform Research Project of Wuhan University Engineering Management Project Center.

\bibliographystyle{IEEEtran}
\bibliography{reference}

@String{Computer = "{IEEE} Computer" }

@String{Springer = "Springer-Verlag" }

@inproceedings{diskt,
  author       = {Yiyun, Zhou and
                  Zheqi, Lv and
                  Shengyu, Zhang and
                  Jingyuan, Chen},
  title        = {Disentangled Knowledge Tracing for Alleviating Cognitive Bias},
  booktitle    = {Proceedings of the {ACM} on Web Conference},
  pages        = {2633--2645},
  year         = {2025}
}

@article{bkt,
  title={Knowledge tracing: Modeling the acquisition of procedural knowledge},
  author={Corbett, Albert T and Anderson, John R},
  journal={User Modeling and User-Adapted Interaction},
  volume={4},
  number={4},
  pages={253--278},
  year={1994},
  publisher={Springer}
}

@article{irt-review,
  title={Item response theory and clinical measurement},
  author={Reise, Steven P. and Waller, Niels G.},
  journal={Annual Review of Clinical Psychology},
  volume={5},
  number={Volume 5, 2009},
  pages={27--48},
  year={2009},
  publisher={Annual Reviews}
}

@inproceedings{dkt,
  author       = {Chris Piech and
                  Jonathan Bassen and
                  Jonathan Huang and
                  Surya Ganguli and
                  Mehran Sahami and
                  Leonidas J. Guibas and
                  Jascha Sohl{-}Dickstein},
  editor       = {Corinna Cortes and
                  Neil D. Lawrence and
                  Daniel D. Lee and
                  Masashi Sugiyama and
                  Roman Garnett},
  title        = {Deep Knowledge Tracing},
  booktitle    = {Advances in Neural Information Processing Systems 28: Annual Conference
                  on Neural Information Processing Systems 2015, December 7-12, 2015,
                  Montreal, Quebec, Canada},
  pages        = {505--513},
  year         = {2015}
}

@inproceedings{dkvmn,
  title={Dynamic key-value memory networks for knowledge tracing},
  author={Zhang, Jiani and Shi, Xingjian and King, Irwin and Yeung, Dit-Yan},
  booktitle={Proceedings of the 26th international conference on World Wide Web},
  pages={765--774},
  year={2017}
}

@inproceedings{akt,
  title={Context-aware attentive knowledge tracing},
  author={Ghosh, Aritra and Heffernan, Neil and Lan, Andrew S},
  booktitle={Proceedings of the 26th ACM SIGKDD international conference on knowledge discovery \& data mining},
  pages={2330--2339},
  year={2020}
}

@inproceedings{gkt,
  title={Graph-based knowledge tracing: modeling student proficiency using graph neural network},
  author={Nakagawa, Hiromi and Iwasawa, Yusuke and Matsuo, Yutaka},
  booktitle={IEEE/WIC/ACM international conference on web intelligence},
  pages={156--163},
  year={2019}
}

@inproceedings{dtransformer,
  title={Tracing knowledge instead of patterns: Stable knowledge tracing with diagnostic transformer},
  author={Yin, Yu and Dai, Le and Huang, Zhenya and Shen, Shuanghong and Wang, Fei and Liu, Qi and Chen, Enhong and Li, Xin},
  booktitle={Proceedings of the ACM web conference 2023},
  pages={855--864},
  year={2023}
}

@inproceedings{cl4kt,
  title={Contrastive learning for knowledge tracing},
  author={Lee, Wonsung and Chun, Jaeyoon and Lee, Youngmin and Park, Kyoungsoo and Park, Sungrae},
  booktitle={Proceedings of the ACM web conference 2022},
  pages={2330--2338},
  year={2022}
}

@inproceedings{dice,
  title={Disentangling user interest and conformity for recommendation with causal embedding},
  author={Zheng, Yu and Gao, Chen and Li, Xiang and He, Xiangnan and Li, Yong and Jin, Depeng},
  booktitle={Proceedings of the web conference 2021},
  pages={2980--2991},
  year={2021}
}

@inproceedings{decrs,
  title={Deconfounded recommendation for alleviating bias amplification},
  author={Wang, Wenjie and Feng, Fuli and He, Xiangnan and Wang, Xiang and Chua, Tat-Seng},
  booktitle={Proceedings of the 27th ACM SIGKDD conference on knowledge discovery \& data mining},
  pages={1717--1725},
  year={2021}
}

@inproceedings{cikt,
  title={Cikt: Causality inspired knowledge tracing},
  author={Zu, Shuaishuai and Li, Li and Cai, Songtao and Shen, Jun},
  booktitle={International Conference on Database Systems for Advanced Applications},
  pages={485--495},
  year={2024},
  organization={Springer}
}

@article{causalKT,
  author       = {Jia Zhu and
                  Xiaodong Ma and
                  Changqin Huang},
  title        = {Stable Knowledge Tracing Using Causal Inference},
  journal      = {{IEEE} Trans. Learn. Technol.},
  volume       = {17},
  pages        = {124--134},
  year         = {2024},
}

@article{tcnet,
  author       = {Changqin Huang and
                  Hangjie Wei and
                  Qionghao Huang and
                  Fan Jiang and
                  Zhongmei Han and
                  Xiaodi Huang},
  title        = {Learning consistent representations with temporal and causal enhancement for knowledge tracing},
  journal      = {Expert Syst. Appl.},
  volume       = {245},
  pages        = {123128},
  year         = {2024}
}

@article{assist,
  author       = {Mingyu Feng and
                  Neil T. Heffernan and
                  Kenneth R. Koedinger},
  title        = {Addressing the assessment challenge with an online system that tutors
                  as it assesses},
  journal      = {User Model. User Adapt. Interact.},
  volume       = {19},
  number       = {3},
  pages        = {243--266},
  year         = {2009},
}

@inproceedings{ednet,
  author       = {Youngduck Choi and
                  Youngnam Lee and
                  Dongmin Shin and
                  Junghyun Cho and
                  Seoyon Park and
                  Seewoo Lee and
                  Jineon Baek and
                  Chan Bae and
                  Byungsoo Kim and
                  Jaewe Heo},
  editor       = {Ig Ibert Bittencourt and
                  Mutlu Cukurova and
                  Kasia Muldner and
                  Rose Luckin and
                  Eva Mill{\'{a}}n},
  title        = {EdNet: {A} Large-Scale Hierarchical Dataset in Education},
  booktitle    = {Artificial Intelligence in Education - 21st International Conference,
                  {AIED} 2020, Ifrane, Morocco, July 6-10, 2020, Proceedings, Part {II}},
  series       = {Lecture Notes in Computer Science},
  volume       = {12164},
  pages        = {69--73},
  publisher    = {Springer},
  year         = {2020}
}

@inproceedings{spanish,
  author       = {Robert V. Lindsey and
                  Mohammad Khajah and
                  Michael C. Mozer},
  editor       = {Zoubin Ghahramani and
                  Max Welling and
                  Corinna Cortes and
                  Neil D. Lawrence and
                  Kilian Q. Weinberger},
  title        = {Automatic Discovery of Cognitive Skills to Improve the Prediction
                  of Student Learning},
  booktitle    = {Advances in Neural Information Processing Systems 27: Annual Conference
                  on Neural Information Processing Systems 2014, December 8-13 2014,
                  Montreal, Quebec, Canada},
  pages        = {1386--1394},
  year         = {2014}
}

@article{slepemapy,
  author       = {Jan Papousek and
                  Radek Pel{\'{a}}nek and
                  V{\'{\i}}t Stanislav},
  title        = {Adaptive Geography Practice Data Set},
  journal      = {J. Learn. Anal.},
  volume       = {3},
  number       = {2},
  pages        = {317--321},
  year         = {2016}
}

@inproceedings{sakt,
  author       = {Shalini Pandey and
                  George Karypis},
  editor       = {Michel C. Desmarais and
                  Collin F. Lynch and
                  Agathe Merceron and
                  Roger Nkambou},
  title        = {A Self Attentive model for Knowledge Tracing},
  booktitle    = {Proceedings of the 12th International Conference on Educational Data
                  Mining, {EDM} 2019, Montr{\'{e}}al, Canada, July 2-5, 2019},
  publisher    = {International Educational Data Mining Society {(IEDMS)}},
  year         = {2019}
}

@inproceedings{sparsekt,
  author       = {Shuyan Huang and
                  Zitao Liu and
                  Xiangyu Zhao and
                  Weiqi Luo and
                  Jian Weng},
  editor       = {Hsin{-}Hsi Chen and
                  Wei{-}Jou (Edward) Duh and
                  Hen{-}Hsen Huang and
                  Makoto P. Kato and
                  Josiane Mothe and
                  Barbara Poblete},
  title        = {Towards Robust Knowledge Tracing Models via k-Sparse Attention},
  booktitle    = {Proceedings of the 46th International {ACM} {SIGIR} Conference on
                  Research and Development in Information Retrieval, {SIGIR} 2023, Taipei,
                  Taiwan, July 23-27, 2023},
  pages        = {2441--2445},
  publisher    = {{ACM}},
  year         = {2023}
}

@inproceedings{simplekt,
  author       = {Zitao Liu and
                  Qiongqiong Liu and
                  Jiahao Chen and
                  Shuyan Huang and
                  Weiqi Luo},
  title        = {simpleKT: {A} Simple But Tough-to-Beat Baseline for Knowledge Tracing},
  booktitle    = {The Eleventh International Conference on Learning Representations,
                  {ICLR} 2023, Kigali, Rwanda, May 1-5, 2023},
  publisher    = {OpenReview.net},
  year         = {2023}
}

@misc{algebra05,
  author       = {John Stamper and Alexandru Niculescu-Mizil and Steven Ritter and Geoffrey Gordon and Kenneth R. Koedinger},
  title        = {{Algebra I 2005-2006 and Bridge to Algebra 2006-2007. Development data sets from {KDD Cup} 2010 Educational Data Mining Challenge}},
  year         = {2010}
}

@inproceedings{deepirt,
  author       = {Chun{-}Kit Yeung},
  editor       = {Michel C. Desmarais and
                  Collin F. Lynch and
                  Agathe Merceron and
                  Roger Nkambou},
  title        = {Deep-IRT: Make Deep Learning Based Knowledge Tracing Explainable Using
                  Item Response Theory},
  booktitle    = {Proceedings of the 12th International Conference on Educational Data
                  Mining, {EDM} 2019, Montr{\'{e}}al, Canada, July 2-5, 2019},
  publisher    = {International Educational Data Mining Society {(IEDMS)}},
  year         = {2019}
}

@inproceedings{saint,
  author       = {Youngduck Choi and
                  Youngnam Lee and
                  Junghyun Cho and
                  Jineon Baek and
                  Byungsoo Kim and
                  Yeongmin Cha and
                  Dongmin Shin and
                  Chan Bae and
                  Jaewe Heo},
  editor       = {David A. Joyner and
                  Ren{\'{e}} F. Kizilcec and
                  Susan Singer},
  title        = {Towards an Appropriate Query, Key, and Value Computation for Knowledge
                  Tracing},
  booktitle    = {L@S'20: Seventh {ACM} Conference on Learning @ Scale, Virtual Event,
                  USA, August 12-14, 2020},
  pages        = {341--344},
  publisher    = {{ACM}},
  year         = {2020}
}

@inproceedings{sinkt,
  author       = {Lingyue Fu and
                  Hao Guan and
                  Kounianhua Du and
                  Jianghao Lin and
                  Wei Xia and
                  Weinan Zhang and
                  Ruiming Tang and
                  Yasheng Wang and
                  Yong Yu},
  editor       = {Edoardo Serra and
                  Francesca Spezzano},
  title        = {{SINKT:} {A} Structure-Aware Inductive Knowledge Tracing Model with
                  Large Language Model},
  booktitle    = {Proceedings of the 33rd {ACM} International Conference on Information
                  and Knowledge Management, {CIKM} 2024, Boise, ID, USA, October 21-25,
                  2024},
  pages        = {632--642},
  publisher    = {{ACM}},
  year         = {2024}
}

@inproceedings{mikt,
  author       = {Jianwen Sun and
                  Fenghua Yu and
                  Qian Wan and
                  Qing Li and
                  Sannyuya Liu and
                  Xiaoxuan Shen},
  editor       = {Tat{-}Seng Chua and
                  Chong{-}Wah Ngo and
                  Ravi Kumar and
                  Hady W. Lauw and
                  Roy Ka{-}Wei Lee},
  title        = {Interpretable Knowledge Tracing with Multiscale State Representation},
  booktitle    = {Proceedings of the {ACM} on Web Conference 2024, {WWW} 2024, Singapore,
                  May 13-17, 2024},
  pages        = {3265--3276},
  publisher    = {{ACM}},
  year         = {2024}
}

@article{corekt,
  author       = {Chaoran Cui and
                  Hebo Ma and
                  Chen Zhang and
                  Chunyun Zhang and
                  Yumo Yao and
                  Meng Chen and
                  Yuling Ma},
  title        = {Do We Fully Understand Students' Knowledge States? Identifying and
                  Mitigating Answer Bias in Knowledge Tracing},
  journal      = {CoRR},
  volume       = {abs/2308.07779},
  year         = {2023}
}

@inproceedings{transformer,
  author       = {Xiaoyang Lu and
                  Boyu Long and
                  Xiaoming Chen and
                  Yinhe Han and
                  Xian{-}He Sun},
  editor       = {Benjamin C. Lee and
                  Harry Xu and
                  Mark Silberstein and
                  Bingyao Li},
  title        = {{I/O} Analysis is All You Need: An {I/O} Analysis for Long-Sequence
                  Attention},
  booktitle    = {Proceedings of the 31st {ACM} International Conference on Architectural
                  Support for Programming Languages and Operating Systems, Volume 2,
                  {ASPLOS} 2026, Pittsburgh, PA, USA, March 22-26, 2026},
  pages        = {962--977},
  publisher    = {{ACM}},
  year         = {2026},
}

@article{dropout,
  author       = {Nitish Srivastava and
                  Geoffrey E. Hinton and
                  Alex Krizhevsky and
                  Ilya Sutskever and
                  Ruslan Salakhutdinov},
  title        = {Dropout: a simple way to prevent neural networks from overfitting},
  journal      = {J. Mach. Learn. Res.},
  volume       = {15},
  number       = {1},
  pages        = {1929--1958},
  year         = {2014}
}

@book{human_performance,
  author    = {Fitts, Paul Morris and Posner, Michael I.},
  title     = {Human Performance},
  publisher = {Brooks/Cole Publishing Company},
  year      = {1967},
  address   = {Belmont, CA, USA},
  note      = {Includes bibliographical references (pages 151-158)},
  lccn      = {67011662}
}

@article{lstm,
  author       = {Sepp Hochreiter and
                  J{\"{u}}rgen Schmidhuber},
  title        = {Long Short-Term Memory},
  journal      = {Neural Comput.},
  volume       = {9},
  number       = {8},
  pages        = {1735--1780},
  year         = {1997},
}

@inproceedings{aiapp,
  title={A Case Study on The Application of Artificial Intelligence in Education Industry},
  author={Qingyun Tang},
  year={2023},
  booktitle={Proceedings of the 2023 3rd International Conference on Modern Educational Technology and Social Sciences (ICMETSS 2023)},
  pages={99-109},
  issn={2352-5398},
  publisher={Atlantis Press}
}

@article{xkt,
  author       = {Chang{-}Qin Huang and
                  Qionghao Huang and
                  Xiaodi Huang and
                  Hua Wang and
                  Ming Li and
                  Kwei{-}Jay Lin and
                  Yi Chang},
  title        = {{XKT:} Toward Explainable Knowledge Tracing Model With Cognitive Learning
                  Theories for Questions of Multiple Knowledge Concepts},
  journal      = {{IEEE} Trans. Knowl. Data Eng.},
  volume       = {36},
  number       = {11},
  pages        = {7308--7325},
  year         = {2024},
}

@article{disenkt,
  author       = {Jiawei Li and
                  Shun Mao and
                  Yixiu Qin and
                  Feng Wang and
                  Yuncheng Jiang},
  title        = {Hyperbolic Hypergraph Transformer With Knowledge State Disentanglement
                  for Knowledge Tracing},
  journal      = {{IEEE} Trans. Knowl. Data Eng.},
  volume       = {37},
  number       = {8},
  pages        = {4677--4690},
  year         = {2025},
}

@inproceedings{grkt,
  author       = {Jiajun Cui and
                  Hong Qian and
                  Bo Jiang and
                  Wei Zhang},
  editor       = {Ricardo Baeza{-}Yates and
                  Francesco Bonchi},
  title        = {Leveraging Pedagogical Theories to Understand Student Learning Process
                  with Graph-based Reasonable Knowledge Tracing},
  booktitle    = {Proceedings of the 30th {ACM} {SIGKDD} Conference on Knowledge Discovery
                  and Data Mining, {KDD} 2024, Barcelona, Spain, August 25-29, 2024},
  pages        = {502--513},
  publisher    = {{ACM}},
  year         = {2024}
}

@article{ekt,
  author       = {Qi Liu and
                  Zhenya Huang and
                  Yu Yin and
                  Enhong Chen and
                  Hui Xiong and
                  Yu Su and
                  Guoping Hu},
  title        = {{EKT:} Exercise-Aware Knowledge Tracing for Student Performance Prediction},
  journal      = {{IEEE} Trans. Knowl. Data Eng.},
  volume       = {33},
  number       = {1},
  pages        = {100--115},
  year         = {2021},
}

@book{irt,
  title     = {Applications of Item Response Theory to Practical Testing Problems},
  author    = {Lord, F. M.},
  year      = {1980},
  publisher = {Routledge},
  address   = {New York},
  edition   = {1st},
  doi       = {10.4324/9780203056615},
  isbn      = {9780203056615},
  pages     = {288}
}

@article{survey2023,
  author       = {Ghodai Abdelrahman and
                  Qing Wang and
                  Bernardo Pereira Nunes},
  title        = {Knowledge Tracing: {A} Survey},
  journal      = {{ACM} Comput. Surv.},
  volume       = {55},
  number       = {11},
  pages        = {224:1--224:37},
  year         = {2023},
}

@article{survey2024,
  author       = {Shuanghong Shen and
                  Qi Liu and
                  Zhenya Huang and
                  Yonghe Zheng and
                  Minghao Yin and
                  Minjuan Wang and
                  Enhong Chen},
  title        = {A Survey of Knowledge Tracing: Models, Variants, and Applications},
  journal      = {{IEEE} Trans. Learn. Technol.},
  volume       = {17},
  pages        = {1898--1919},
  year         = {2024},
}

@article{tlsbkt,
  author       = {Kai Zhang and
                  Yiyu Yao},
  title        = {A three learning states Bayesian knowledge tracing model},
  journal      = {Knowl. Based Syst.},
  volume       = {148},
  pages        = {189--201},
  year         = {2018},
}

@inproceedings{rckt,
  author       = {Jiajun Cui and
                  Minghe Yu and
                  Bo Jiang and
                  Aimin Zhou and
                  Jianyong Wang and
                  Wei Zhang},
  title        = {Interpretable Knowledge Tracing via Response Influence-based Counterfactual
                  Reasoning},
  booktitle    = {40th {IEEE} International Conference on Data Engineering, {ICDE} 2024,
                  Utrecht, The Netherlands, May 13-16, 2024},
  pages        = {1103--1116},
  publisher    = {{IEEE}},
  year         = {2024},
}



\vfill

\vfill

\end{document}